# An End-to-End Deep Reinforcement Learning Approach for Solving the Traveling Salesman Problem with Drones

Taihelong Zeng [1], Yun Lin [1, *], Yuhe Shi [2], Yan Li [1, 3], Zhiqing Wei [1], Xuanru Ji [4]

1. School of Management Science and Real Estate, Chongqing University, Chongqing, 400045, PR China
2. School of Management, Guizhou University, Guiyang, 550025, PR China
3. Chongqing Changan Minsheng APLL Logistics Co., Ltd., Chongqing, 401122, PR China
4. State Grid Chongqing Electric Power Company Material Branch, Chongqing, 401120, PR China

- Corresponding author: Dr. Yun Lin, Email: linyun@cqu.edu.cn

Homepage: https://faculty.cqu.edu.cn/YunLin/zh_CN/index.htm

http://www.msre.cqu.edu.cn/szdw/jslb/ly2.htm

# Abstract

The emergence of truck-drone collaborative systems in last-mile logistics has positioned the Traveling Salesman Problem with Drones (TSP-D) as a pivotal extension of classical routing optimization, where synchronized vehicle coordination promises substantial operational efficiency and reduced environmental impact, yet introduces NP-hard combinatorial complexity beyond the reach of conventional optimization paradigms. Deep reinforcement learning offers a theoretically grounded framework to address TSP-D's inherent challenges through self-supervised policy learning and adaptive decision-making. This study proposes a hierarchical Actor-Critic deep reinforcement learning framework for solving the TSP-D problem. The architecture consists of two primary components: a Transformer-inspired encoder and an efficient Minimal Gated Unit decoder. The encoder incorporates a novel, optimized k-nearest neighbors sparse attention mechanism specifically for focusing on relevant spatial relationships, further enhanced by the integration of global node features. The Minimal Gated Unit decoder processes these encoded representations to efficiently generate solution sequences. The entire framework operates within an asynchronous advantage actor-critic paradigm. Experimental results show that, on benchmark TSP-D instances of various scales (N=10 to 100), the proposed model can obtain competitive or even superior solutions in shorter average computation times compared to high-performance heuristic algorithms and existing reinforcement learning methods. Moreover, compared to advanced reinforcement learning algorithm benchmarks, the proposed framework significantly reduces the total training time required while achieving superior final performance, highlighting its notable advantage in training efficiency.

# Keywords

# 1. Introduction

Contemporary logistics systems, especially in last-mile delivery operations, increasingly confront challenges related to costs, operational efficiency, and environmental sustainability (Pourmohammadreza et al., 2025). In this context, Unmanned Aerial Vehicles (UAVs) or drones have emerged as promising solutions due to their operational flexibility and ability to navigate around ground-based obstacles (Li et al., 2022). Collaborative delivery models combining drones with traditional trucks have consequently emerged, among which the Traveling Salesman Problem with Drones (TSP-D) has garnered significant attention as a representative paradigm (Murray and Chu, 2015; Zhou et al., 2025). This paradigm leverages the large payload capacity and long range of trucks combined with the efficient short-haul delivery capabilities of drones, aiming to significantly reduce delivery times, lower operational costs, and improve delivery coverage and responsiveness in complex environments (such as urban or remote areas), demonstrating significant theoretical research value and practical application prospects (Lu et al., 2025).

The TSP-D problem can be described as a hybrid delivery system where one or more trucks, carrying several drones, depart from a distribution center to serve a set of customer locations (Wang et al., 2023). During the delivery process, drones can be launched from truck stopping points to independently serve nearby customers before rendezvousing with the truck at predetermined locations (Murray and Chu, 2015). TSP-D is an extension of the classic Traveling Salesman Problem (TSP) and Vehicle Routing Problem (VRP). However, unlike traditional VRP, TSP-D introduces collaborative decision-making between trucks and drones, task allocation, path planning, and precise time synchronization requirements, which significantly increases the problem's complexity (Duan et al., 2025). The solution space expands dramatically with the number of customers, drones, and associated constraints, and the problem has been proven to be NP-hard, posing severe challenges to the efficiency and quality of solution algorithms (Yu et al., 2023).

For combinatorial optimization problems like TSP-D, traditional solution approaches are primarily categorized into two classes: exact algorithms and heuristic algorithms. Exact algorithms including branch-and-bound, dynamic programming, and integer linear programming guarantee global optimal solutions but suffer from exponential computational complexity growth in large-scale instances, making them impractical for real-world applications (Laporte, 1992). Heuristic algorithms like simulated annealing, genetic algorithms, and ant colony optimization achieve computational efficiency at the expense of solution quality, gaining widespread adoption in practical implementations. However, these methods often require manual design for specific problem instances, exhibit poor adaptability, and risk converging to local optima (Karimi-Mamaghan et al., 2022). Recently, Deep Reinforcement Learning (DRL) has emerged as a cutting-edge AI technology demonstrating significant potential in solving combinatorial optimization problems through its powerful representation learning and decision optimization capabilities (Bengio et al., 2021). Compared with traditional methods, DRL approaches offer advantages including end-to-end learning, enhanced generalization capabilities, and elimination of manual heuristic design. Growing evidence demonstrates their superior performance over conventional heuristics in solving large-scale TSP problems (Liu et al., 2022; Zhao et al., 2021), establishing a new paradigm for path planning solutions.

While deep reinforcement learning has made remarkable progress in solving conventional TSP and VRP problems, research specifically addressing TSP-D remains relatively scarce, with existing studies predominantly limited to single-agent perspectives. The challenges of TSP-D primarily manifest in three aspects: (1) Dual-agent coordination requiring simultaneous optimization of truck and drone routes; (2) The speed difference between trucks and drones complicates the calculation of travel time; (3) Spatiotemporal synchronization requiring precise coordination of truck-drone rendezvous at specified locations and timestamps (Agatz et al., 2018).

Therefore, we propose an Actor-Critic-based deep reinforcement learning framework specifically designed for TSP-D resolution. The design of this solution involves several crucial components:

(1) We propose a Transformer-inspired encoder architecture. By integrating enhanced k-nearest neighbors sparse attention mechanisms with dynamic graph masking and global node features, this encoder effectively captures the crucial spatial structural patterns inherent in TSP-D problems while significantly reducing computational complexity.

(2) We replace conventional Gated Recurrent Units (GRU) / Long Short-Term Memory (LSTM) structures with an efficient Minimal Gated Unit (MGU) to enhancing the encoder's efficiency. This substitution maintains the model's representational capacity for sequential decision-making while substantially reducing the parameter count and accelerating inference speed, contributing to a leaner overall model.

(3) For effectively training process, we employ an asynchronous advantage actor-critic framework. This learning process is further enhanced through prioritized experience replay and adaptive learning rate scheduling, strategically improving sample efficiency and overall training stability.

(4) Building upon the adaptive training mechanisms, we introduce a reward plateau-aware adaptive learning rate strategy. This specifically targets and resolves convergence challenges often encountered in large-scale TSP-D instances.

The remainder of this paper is organized as follows. Section 2 briefly reviews current research on TSP-D path optimization, introducing both traditional methods and deep reinforcement learning approaches. Section 3 formally defines the TSP-D problem and formulates it through Markov Decision Processes. Section 4 presents the proposed Actor-Critic framework and training strategies. Section V details computational experiments and analytical results. Finally, Section 5 concludes the paper and outlines future research directions.

# 2 Literature Review

2.1. Traveling Salesman Problem with Drones

Since (Murray and Chu, 2015) first proposed the vehicle-drone collaborative delivery problem, the TSP-D has gained significant attention in recent years as an optimization method for delivery systems. Current research identifies two primary collaborative delivery models integrating vehicles and drones. The first model employs independent delivery operations, where vehicles and drones depart from distribution centers with cargo, complete their respective delivery tasks, and subsequently return to the centers. (Ham, 2018) addressed this model by proposing a multi-vehicle, multi-drone, multi-depot pickup-delivery problem, where drones can collect goods at subsequent nodes after deliveries before returning to depots, optimizing independent vehicle-drone routes for global efficiency. (Zhang et al., 2023) established a mixed-integer programming model for electric vehicle-drone collaborative delivery systems with time windows and endurance constraints, and developed an extended adaptive large neighborhood search algorithm for efficient solution generation. (Nguyen et al., 2022) restricted drone usage by defining maximum operation time constraints, extending the problem with parallel drone scheduling. The second category features collaborative delivery operations, where drones are carried by vehicles departing from depots, deployed at specific nodes, and later rendezvous with vehicles after completing deliveries. (Wang et al., 2017) first established mixed-integer linear programming formulations for dual-drone delivery problems. (Poikonen et al., 2017) extended (Wang et al., 2017)'s work by analyzing worst-case scenarios under battery constraints, incorporating crow-flight distance metrics for drones versus truck routing, and considering transportation costs beyond temporal optimization. (Schermer et al., 2018) comparatively evaluated two novel heuristic approaches: Two-Phase Heuristic (TPH) and Single-Phase Heuristic (SPH). Their work first produced numerical results for large-scale instances, demonstrating TPH's superior performance over SPH in most scenarios. (Sacramento et al., 2019) developed an adaptive large neighborhood search metaheuristic incorporating time-window constraints. (Kitjacharoenchai et al., 2019) addressed operational constraints related to drone launch windows and delivery time coordination. (Mahmoudinazlou and Kwon, 2024) hybridized genetic algorithms with dynamic programming to create the escape strategy-based HGA-TAC$^+$ algorithm for TSP-D resolution, (Wang et al., 2022) formulated a time-dependent truck-drone hybrid routing problem accounting for traffic conditions, and proposed an iterated local search algorithm. A key limitation of the existing drone-assisted truck delivery solutions lies in their reliance on the simplified assumption of single-customer service per drone sortie. To address this constraint and enhance practicality, (Gu et al., 2022) developed a novel approach enabling multi-customer service per drone mission. (Nguyen et al., 2022) incorporated constraints on total truck payload capacity and combined operational durations for both trucks and drones. (Montaña et al., 2021) quantified carbon emissions from last-mile drone deliveries, emphasizing the sustainability implications. (Rave et al., 2023) proposed a novel micro-depot concept. In the problem they defined, drones can be launched from trucks or from microdepots on trucks. (Sitek et al., 2022) defined this problem as the Extended Vehicle Routing Problem with Drones, which differs from the traditional Vehicle Routing Problem with Drones in that they consider what are called mobile hubs (similar to microdepots) (Rave et al., 2023) as mobile points from which drones can be launched. At the same time, a genetic algorithm was developed to minimize operational costs and select mobile hubs. (Wu et al., 2022) investigated the effects of payload and flight duration on drone energy consumption, creating a variable neighborhood descent-based heuristic algorithm.

2.2. Deep reinforcement learning solves path planning

With the rapid advancement of artificial intelligence, Deep Reinforcement Learning, integrating deep learning's representational power with reinforcement learning's decision-making capabilities, has demonstrated significant potential in solving combinatorial optimization problems, particularly in path planning. Ranging from the initial exploration of (Hopfield and Tank, 1985) neural networks to the widespread application of deep learning in recent years, DRL-based path planning research has achieved remarkable progress. Early research primarily focused on employing neural networks to directly solve small-scale TSP and VRP. The Hopfield neural network represents one of the earliest attempts, utilizing energy function concepts to derive approximate solutions. The subsequent development of Pointer Networks (PN) marked a new phase in applying deep learning techniques to path planning challenges. (Sutskever et al., 2014) enhanced the Seq2Seq framework to develop the PN, which employs LSTM-based encoding and attention mechanism decoding to effectively extract node coordinate features. This work laid the foundation for applying deep reinforcement learning to path planning challenges, However, this supervised training paradigm required extensive TSP datasets with optimal tours, limiting its practicality. This methodology proved challenging to implement, as the supervised training process inherently constrained solution quality to the performance level of the training data. To address these limitations, (Bello et al., 2016) implemented an Actor-Critic DRL training algorithm to optimize PN without requiring pre-computed optimal tours. This demonstrates the feasibility of DRL in learning high-quality strategies without supervised labels, and establishes an important foundational framework for numerous subsequent DRL-based path planning studies, including our current work. (Nazari et al., 2018) observed that the RNN structure in the original PN might introduce unnecessary complexity. They simplified the architecture, directly used embedded inputs, and incorporated dynamic element inputs, proposing the NM model. For multi-objective planning problems, (Li et al., 2021) adopted an improved PN similar to (Nazari et al., 2018) for modeling. Using an RNN model based on GRU (Cho et al., 2014) , this model has similar performance to the LSTM used in the original PN in (Nazari et al., 2018), but with fewer parameters. (Kong and Jiang, 2024) designed a composite PN framework integrating graph neural network representations and attention mechanisms to address Drone-Assisted truck transportation. (Shi, 2022) integrated PN with heuristic algorithms like Variable Neighborhood Search to enhance initial solution quality in Traveling Salesman Problem applications. Experimental results demonstrated improved convergence rates and optimization outcomes. The proposed method exhibited superior generalization capability and faster training speeds compared to PN networks on TSP and Capacitated Vehicle Routing Problem (CVRP) benchmarks. Despite the relative success of PN and their variants, they still show limitations in capturing complex dependencies between nodes, especially in large-scale problems, which has prompted researchers to explore more powerful architectures.

Following Google's introduction of the Transformer architecture combining attention mechanisms and multilayer perceptron (Vaswani et al., 2017), attention-based approaches have proliferated in this problem domain. (Kool et al., 2018) first adapted Transformer architectures for route optimization, developing an Attention model (AM) framework that significantly improved both training efficiency and solution optimality for various path planning tasks, The AM model became an important benchmark for subsequent research, but it completely removed positional encoding in the encoder, which may result in the loss of important spatial information. Additionally, the self-attention mechanism of standard Transformers faces a bottleneck of quadratic computational complexity when processing large-scale graphs. Building on (Kool et al., 2018) 's experimental framework, (Joshi et al., 2019) combined supervised learning and reinforcement learning (RL) to train models on 100-node TSP instances, achieving improved prediction accuracy. (Peng et al., 2019) enhanced the AM framework with dynamic decoding mechanisms, achieving superior solution accuracy and generalization capabilities for CVRP problems, (Zhang et al., 2020) developed a Multi-Agent Attention Model (MAAM) encoder-decoder framework using Transformer architectures, featuring attention layers that iteratively generate multi-vehicle routes. (Xu et al., 2022) advanced AM research by incorporating dynamic network architectures through attention aggregation modules across multiple relational structures, enabling context-aware embeddings that enhance model performance and generalization. These works demonstrate the powerful potential of the Transformer architecture in path planning, but also highlight the need to further enhance its representational capacity and efficiency, especially in scenarios requiring the capture of complex graph structures and long-distance dependencies.

Researchers have also explored combining Transformer with other technologies or modifying it. (Deudon et al., 2018) replaced the recurrent encoder in the PN framework with Transformer-based attention encoding for TSP heuristic learning. Integration with 2-opt heuristics further enhanced solution quality. (Ma et al., 2019) proposed a Dual-Aspect Collaborative Transformer (DACT) with cyclic positional encoding (CPE) to separately learn node and positional embeddings, applying it to TSP and CVRP. (Gebreyesus et al., 2023) implemented a Transformer-based encoder-decoder architecture with Edge-Enhanced Multi-Head Attention (EEMHA) layers in the encoder.

In terms of decoders, those based on recurrent networks (especially LSTM) have been widely applied due to their ability to maintain state information during the sequential decision-making process (Nazari et al., 2018; Peng et al., 2019). For the TSP-D problem, the relative positions and coordination between vehicles in a heterogeneous fleet are crucial, and the hidden state memory capability of LSTM may be more advantageous than stateless attention decoders. As (Bogyrbayeva et al., 2023) innovatively integrated AM and NM, adopted an attention-based encoder and LSTM decoder in their model for TSP-D. However, LSTM has a relatively large number of parameters. This inspires us to consider whether there exists a more lightweight recurrent unit that can both capture the temporal dynamics needed for TSP-D coordination and maintain high efficiency.

Through a review and analysis of existing literature, we found that:

(1) Actor-Critic framework has proven to be an effective DRL paradigm for solving path planning problems.

(2) Attention models show greater potential in encoding graph structure information compared to earlier PN and its variants, but standard attention mechanisms have limitations in position information encoding and computational efficiency for large-scale graphs.

(3) The choice of decoder is important for sequential decision tasks. For heterogeneous agent systems requiring coordination (such as TSP-D), the state memory capability of recurrent networks may be superior to stateless attention, though efficiency issues need to be considered.

(4) Despite the progress of DRL in problems such as TSP and VRP, research on heterogeneous agent systems, especially the drone-assisted TSP-D problem, remains relatively scarce, with existing methods mostly limited to single-agent scenarios.

The trade-off rule between solution quality and runtime in existing algorithms for TSP-D solutions is unavoidable. To address this issue as much as possible, based on the opportunities and shortcomings identified in the above literature analysis, we designed an end-to-end deep reinforcement learning architecture.

# 3 Problem description

## 3.1. Problem Definition

This study focuses on the TSP-D, where a truck and a drone collaboratively execute delivery services for multiple customer nodes. The problem scenario is defined within a network containing n nodes, where $n-1$ nodes represent customers and the remaining node serves as the depot. Each customer node possesses unit demand, while the depot requires no service. Both vehicles depart from the depot, with the objective of minimizing total service time through optimized routing coordination while satisfying all customer demands.

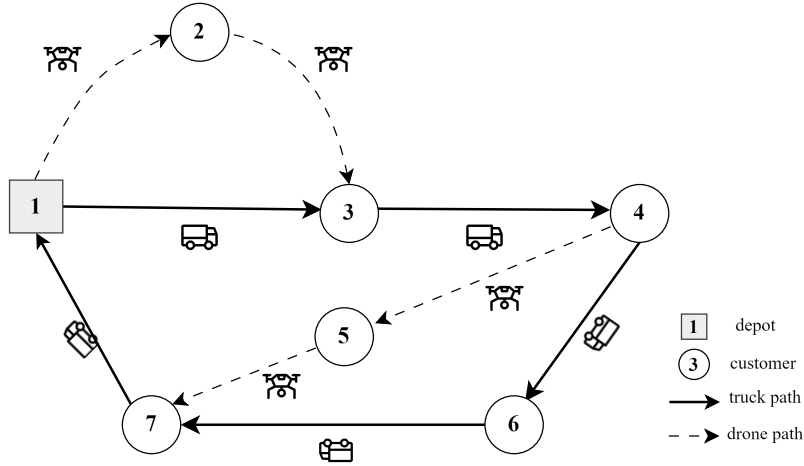

**Fig. 1**. TSP-D problem diagram

As shown in Fig. 1, the truck executes main route delivery tasks along the path 1→3→4→6→7→1 in this distribution scheme. This route forms the backbone of the distribution network, providing multiple launch/recovery points for the drone. The drone services customers deviating from the main route through two distinct flight missions, requiring return to the truck after each delivery. With the truck's dynamic repositioning, the drone may:

(1) rendezvous at the truck's next delivery node (1→2→3).

(2) meet at a subsequent multi-node service location (4→5→7).

(3) remain idle during certain segments (3→4).

## 3.2. MDP-based Formalization

We formulate the TSP-D as a Markov Decision Process (MDP). The MDP provides a mathematical framework for sequential decision-making, where the agent (coordinated truck-drone system) learns optimal policies through environmental interactions.

**(1) State Space**

The state space $S$ encapsulates complete information about truck/drone statuses and customer demands at each decision epoch. A state $s$ can be represented as a tuple:

$$s = \left(\{\boldsymbol{p}_i\}_{i \in N}, \boldsymbol{c}, l_t, l_d, s_d, r_d, \boldsymbol{a}_t, \boldsymbol{a}_d\right) \tag{1}$$

We define the system state $s \in S$ as follows: the environment contains $n$ nodes comprising, $n-1$ customers and one depot. Spatial positions are defined through Cartesian coordinates $(x_i, y_i)$ for each node $i$. The coordinate set is denoted as $\{\boldsymbol{p}_i\}_{i \in N}$, demand vector $\boldsymbol{c} = (c_1, c_2, \ldots, c_{n-1}) \in \{0,1\}^{n-1}$ tracks customer statuses: $c_i = 1$ indicates unmet demand for customer $i$, $c_i = 0$ denotes fulfilled demand. The truck's position $l_t \in N$ and drone's location $l_d \in N$ are continuously tracked. $s_d \in \{0,1\}$ indicates drone operation status: $s_d = 1$ (engaged in delivery), $s_d = 0$ (available). $r_d \in \{0,1\}$ records drone return status: $r_d = 1$ (not returned), $r_d = 0$ (returned). Truck and drone may have pending actions. Let $\boldsymbol{a}_t = (j, \tau_t)$ represent the truck's pending action, where $j \in N$ is the target node and $\tau_t \in R^+$ the remaining time to reach $j$. If no pending action exists, $\boldsymbol{a}_t = (l_t, 0)$. Similarly, define $\boldsymbol{a}_d = (k, \tau_d)$ for drone operations, where $k \in N$ is the drone's target node and, $\tau_d \in R^+$ the remaining flight time. If the drone has no pending action, $\boldsymbol{a}_d = (l_d, 0)$.

The state space is constrained by: In the initial state, all customer demands are unserved $c_i = 1, \forall i \in C$, $C = \{1, 2, \ldots, n-1\}$; The truck and drone initially start at the depot node $l_t = l_d = n$; The drone can only launch when docked with/returned to the truck ($s_d = 0 \Rightarrow l_d = l_t$); When servicing customers, the drone cannot be at the depot ($s_d = 1 \Rightarrow l_d \neq n$).

**(2) Action Space**

The action space $A$ encompasses all feasible actions available to both truck and drone at each decision epoch. Given their collaborative operation, joint actions are represented as $\boldsymbol{a} = (a_t, a_d)$, where $a_t \in N$ denotes the truck's next target node and $a_d \in N$ specifies the drone's subsequent destination.

The action space is constrained by: Trucks and drones cannot visit customer nodes with satisfied demands $a_t, a_d \notin \{i \in C \mid c_i = 0\}$. When the drone is servicing a customer $s_d = 1$, it must return to the truck's current node $a_d = l_t$. If the drone has returned ($r_d = 0$), it cannot relaunch until the truck visits another customer. Vehicles with pending actions ($a_t \neq 0$ or $a_d \neq 0$) must continue executing them until completion.

**(3) Transition Function**

The state transition function $P : S \times A \to S$ specifies the probability of transitioning to state $s$ given current state $s'$ and action $\boldsymbol{a}$. In our problem formulation, state transitions are deterministic: given the current state and action, the subsequent state is uniquely determined. The state transition function can be expressed as $s' = f(s, a)$, where $f$ represents a deterministic mapping.

The state transition process comprises: First, calculate the time step $\Delta t$ required to reach the next state based on vehicle actions and distances to target nodes. Specifically, truck arrival time $t_t = d_{l_t, a_t} / v_t$ and drone arrival time $t_d = d_{l_d, a_d} / v_d$, where $d_{i,j}$ is the distance between nodes $i$ and $j$, $v_t$ and $v_d$ denote respective speeds. Incorporating pending action residuals, $\Delta t = \min(t_t - \tau_t, t_d - \tau_d, \tau_t, \tau_d)$ (excluding terms where $\tau_t$ or $\tau_d = 0$). Update vehicle positions using $\Delta t$ and selected actions: if $\Delta t = t_t - \tau_t$, truck arrives: $l_t' = a_t$; if $\Delta t = t_d - \tau_d$, drone arrives: $l_d' = a_d$. Else positions remain unchanged. Update customer states: if $l_t' \in C$ or $l_d' \in C$, set $c_{l_t'}' = 0$ or $c_{l_d'}' = 0$. Update drone status: if $l_d' \in C$, set service state $s_d' = 1$. If drone returns to truck $(l_d' = l_t')$, set $s_d' = 0$ and $r_d' = 0$. If $s_d$ transitions from $0 \to 1$, set $r_d = 1$, Finally, update pending actions: $a_t' = (a_t, \max(0, \tau_t - \Delta t))$ for truck, $a_d' = (a_d, \max(0, \tau_d - \Delta t))$ for drone.

**(4) Reward Function**

The reward function $R: S \times A \to R$ defines the immediate reward for executing action $a$ in state $s$. To minimize total service duration, we formulate the reward as negative temporal progression:

$$r(s, a) = -\Delta t \tag{2}$$

Where $\Delta t$ represents the time step required to transition from $s$ to $s'$ through action $a$. This design incentivizes the agent to prioritize time-efficient action sequences.

# 4 Solution Model Architecture

To address TSP-D, we propose an end-to-end deep reinforcement learning architecture. Building upon the demonstrated efficacy of Actor-Critic frameworks in solving complex routing problems (Bello et al., 2016; Kool et al., 2018; Nazari et al., 2018), we adopt this paradigm as the foundation of our approach. At the core of our contribution, we develop an Adaptive Expansion Graph Attention for graph representation learning, Our encoder substantially enhances attention mechanisms, employing relative and sparse positional encoding instead of removing them as in AM (Kool et al., 2018) and hybrid model (HM) that uses multi-head attention for encoding (Bogyrbayeva et al., 2023), To overcome computational bottlenecks in traditional Transformers for large-scale graphs and capture long-range dependencies, we adapt Google's Exphormer sparse attention framework (Shirzad et al., 2023), introducing an innovative adaptive graph expansion mechanism. This enables dynamic construction of sparse interaction graphs. Additionally, our encoder incorporates global node mechanisms to enhance contextual awareness. LSTM-based decoders have proven effective for routing problems (Nazari et al., 2018; Peng et al., 2019), For drone-assisted planning where relative vehicle positions in heterogeneous fleets are crucial, LSTM's hidden state memory outperforms stateless attention decoders (Bogyrbayeva et al., 2023). Diverging from HM's LSTM, we investigate MGU decoders as lightweight alternatives, our findings show that MGU achieves comparable generalization to LSTM through simplified gating, sufficiently capturing temporal dynamics essential for TSP-D coordination. As shown in Fig. 2, our model architecture diagram is clearly presented.

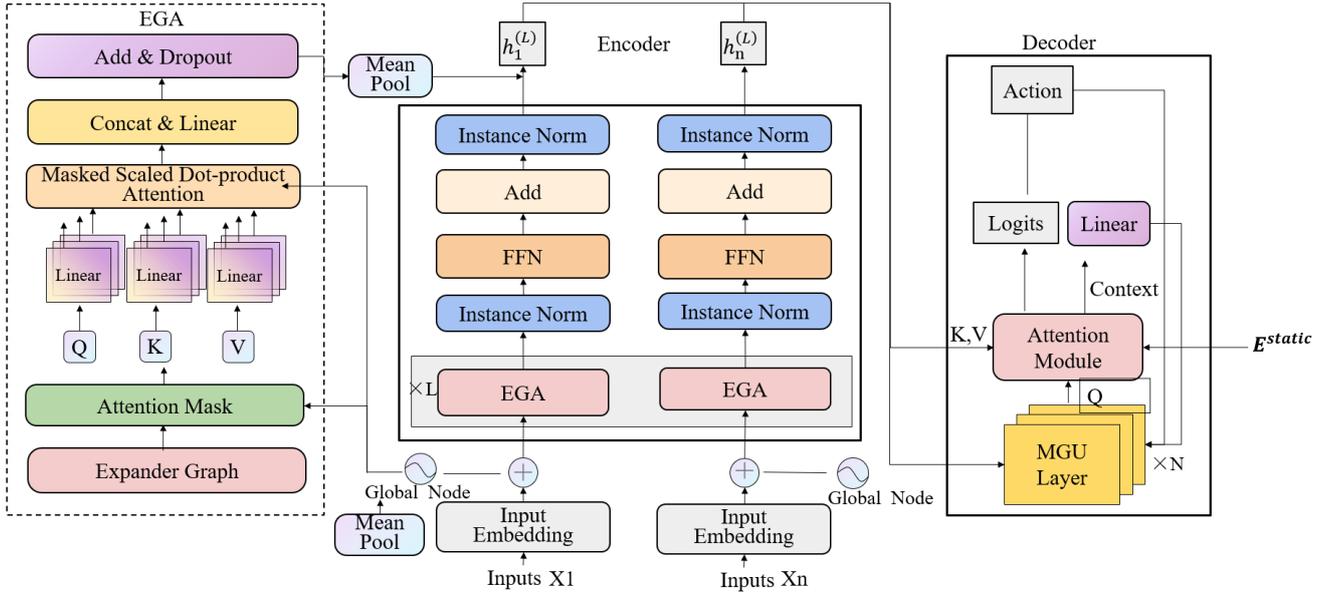

**Fig. 2**. The encoder-decoder framework in our model

## 4.1. Encoder

Given a graph composed of node set $N$ where each node $n \in N$ possesses input features $x_n \in R^{D_{in}}$. To embed this graph structure, we initialize node embeddings $h_n^{(0)}$ through linear transformation:

$$h_n^{(0)} = W_{input} x_n + b_{input} \quad \forall n \in N \tag{3}$$

Where $W_{input} \in R^{D_{in} \times D_h}$ and $b_{input} \in R^{D_h}$ are learnable parameters, with $D_h$ denoting hidden layer dimension. These initial node embeddings $\{h_n^{(0)} : n \in N\}$ are processed through $L$ stacked Expander Graph Attention (EGA) encoder layers.

Each EGA encoder layer dynamically expands the graph structure $G = (N, E)$, where $E$ denotes edge connections. To enhance graph connectivity and incorporate global context, we employ:

(1) k-Nearest Neighbor Connection: Selects the top $k = \lceil \log_2 |N| \rceil$ neighbors based on latent space distance:

$$d_{ij} = \| h_i^{(l-1)} - h_j^{(l-1)} \|_2 \tag{4}$$

(2) Hierarchical Connection: Divides nodes into $L = |N|/8$ layers and establishes connections within each layer.

$$E_{layer} = \bigcup_{l=1}^{L} \{(v_i, v_j) \mid \lfloor i/s \rfloor = \lfloor j/s \rfloor, |i-j| \leq 2\} \tag{5}$$

Where stride $s = |N|/L$.

(3) Depot connections: guarantee direct links between all nodes and a predefined depot node to enable rapid information aggregation and distribution.

Recognizing that expanded graphs may compromise long-range dependency modeling compared to fully-connected graphs, we augment the architecture with a special global node $n_{global}$, extending the adjacency matrix to $\tilde{A} \in \{0,1\}^{(N+1) \times (N+1)}$, this composite matrix integrates original edges, expanded connections, and global node linkages. The relevant schematic diagram is shown in Fig. 3.

To precisely regulate node interactions in attention mechanisms while maintaining structural alignment with $\tilde{A}$, we implement attention mask $M \in \{0, -\infty\}^{(N+1) \times (N+1)}$, each mask element $M_{i,j}$ corresponds respectively with $\tilde{A}$ entries. During attention score calculation. $M$ is additively combined with raw attention scores. This configuration assigns $-\infty$ attention scores to unconnected node pairs, which become zero after Softmax normalization.

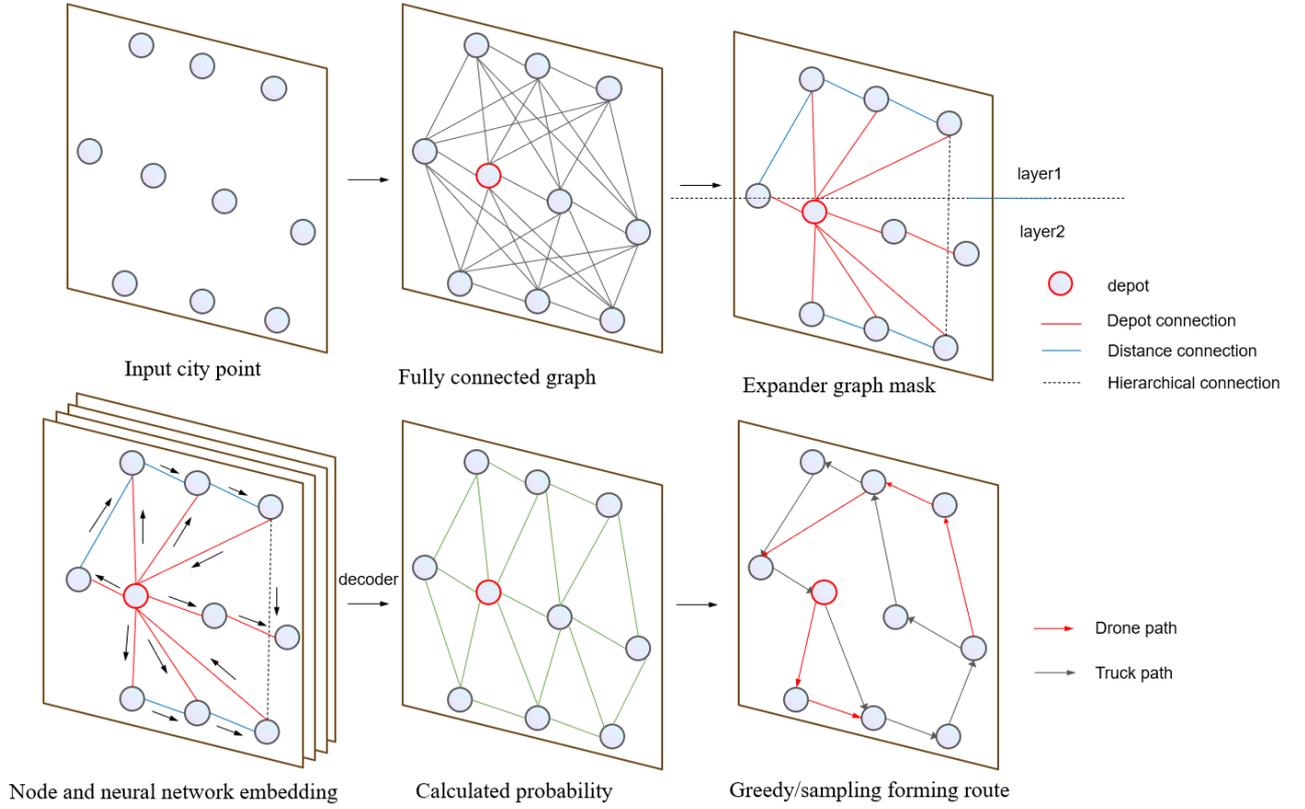

**Fig. 3.** Schematic Diagram of expander graph Masking

For node representations $h_n^{(l-1)}$ input to this sub-layer, we first compute query vectors $q_n \in R^{D_k}$, key vectors $k_n \in R^{D_k}$, and value vectors $v_n \in R^{D_v}$:

$$q_n = W_Q h_n^{(l-1)} \tag{6}$$

$$k_n = W_K h_n^{(l-1)} \tag{7}$$

$$v_n = W_V h_n^{(l-1)}, \ \forall n \in N \tag{8}$$

$W_Q, W_K, W_V$ are learnable weight matrices with dimensions $(D_h \times D_k), (D_h \times D_k)$, and $(D_h \times D_v)$ respectively. To compute attention scores $s_{i,j}$ between nodes $i$ and $j$, we employ masked dot-product attention combined with relative positional encoding:

$$s_{i,j} = q_i^T k_j / \sqrt{D_k} + q_i^T r_{ij} \tag{9}$$

where $r_{ij}$ represents the relative positional encoding vector between nodes $i$ and $j$. In our implementation, relative positional encodings are retrieved from a sparse positional embedding matrix $R \in R^{H \times D_{sparse} \times D_k}$, where $H$ denotes the number of attention heads and $D_{sparse}$ the dimensionality of sparse positional embeddings. Attention weights $a_{i,j}$ are obtained by Softmax normalization of attention scores:

$$a_{i,j} = \exp(s_{i,j}) / \sum_{j'} \exp(s_{i,j'}) \tag{10}$$

The updated representation $h_n'$ for node $n$ is computed as the weighted sum of all nodes'

value vectors:

$$h'_n = \sum_j a_{n,j} v_j \quad (11)$$

To enhance the model's expressive capability, we employ a multi-head attention mechanism utilizing $H$ independent attention heads. The messages computed from each head are concatenated and linearly transformed through an output projection matrix $W_O \in R^{(H \cdot D_v) \times D_h}$ to produce the final EGA sublayer output:

$$\text{EGA}\left(h_1^{(l-1)}, \ldots, h_{N+1}^{(l-1)}\right) = W_O \text{Concat}\left(\left[h_n'^{(1)}, \ldots, h_n'^{(H)}\right]\right) \quad (12)$$

To stabilize the training procedure, we apply Instance Normalization after the EGA sublayer and incorporate residual connections:

$$\hat{h}_n^{(l)} = \text{InstanceNorm}\left(h_n^{(l-1)} + \text{EGA}\left(h_1^{(l-1)}, \ldots, h_{N+1}^{(l-1)}\right)\right) \quad (13)$$

The normalized output $\hat{h}_n^{(l)}$ is subsequently fed into the Feed-Forward Network (FFN) sublayer. The FFN sublayer consists of two linear layers with a ReLU activation function.

$$\text{FF}^{(l)}\left(\hat{h}_n^{(l)}\right) = W_{ff,2} \cdot \text{ReLU}\left(W_{ff,1} \hat{h}_n^{(l)} + b_{ff,1}\right) + b_{ff,2} \quad (14)$$

Mirroring the attention sublayer, we apply Instance Normalization and residual connections after the FFN sublayer:

$$h_n^{(l)} = \text{InstanceNorm}\left(\hat{h}_n^{(l)} + \text{FF}^{(l)}\left(\hat{h}_n^{(l)}\right)\right) \quad (15)$$

At the inception of each EGA layer, we compute mean-pooled features $p^{(l)}$ from original node inputs (excluding the global node):

$$p^{(l)} = \text{Mean}\left(h_{1:N}^{(l-1)}, \text{dim}=1\right) \quad (16)$$

This averaged feature is then added to the global node's representation $h_{N+1}^{(l-1)}$ to enable layer-wise global information updating:

$$h_{N+1}^{(l-1)} = h_{N+1}^{(l-1)} + p^{(l)} \quad (17)$$

After processing through $L$ EGA layers, we obtain final node representations $\left\{h_n^{(L)} : n \in N+1\right\}$. To integrate global information across EGA layers, we compute multi-scale mean-pooled features $p_{multi-scale}$:

$$p_{multi-scale} = \text{Mean}\left(\left[p^{(1)}, \ldots, p^{(L)}\right], \text{dim}=1\right) \quad (18)$$

Finally, we add $p_{multi-scale}$ to the terminal layer's node features $h_n^{(L)}$ to accomplish ultimate information fusion:

$$h_n^{(L)} = h_n^{(L)} + p_{multi-scale} \quad (19)$$

To prepare encoder outputs for decoding, we process the final node features through specific transformations. Specifically, we extract original node features (excluding the global node), apply dimension unification via permute operation, and designate them as static decoder inputs $E^{static}$.

### 4.2. Decoder

These encoded node features will be used to autoregressively generate action sequences for solving combinatorial optimization problems. The decoder employs a unidirectional MGU recurrent neural network integrated with attention mechanisms to select appropriate actions at each decoding step.

The decoder's objective is to progressively construct solution sequences using graph node representations from the encoder and evolving dynamic state information. To achieve this, we implement an L-layer unidirectional MGU recurrent neural network as the decoder's core component. At each decoding timestep $t$, the network receives: static features $E^{static} \in R^{B \times D_h \times N}$, dynamic features $E^{dynamic} \in R^{B \times D_h \times N}$, from the encoder, decoder input $x_{dec}^{(t)} \in R^{B \times D_h}$, and previous hidden state $h^{(t-1)} \in R^{L \times B \times D_h}$. The decoding process proceeds autoregressively until complete solution sequences are generated. The decoder core consists of L stacked MGU recurrent layers. At timestep $t$, the MGU processes decoder input $x_{dec}^{(t)}$ and previous hidden state $h^{(t-1)}$ to update its state, generating current output $r^{(t)} \in R^{B \times D_h}$ and updated hidden state $h^{(t)} \in R^{L \times B \times D_h}$,

The MGU update procedure can be formally expressed as:

$$r^{(t)}, h^{(t)} = \text{MGU}\left(x_{dec}^{(t)}, h^{(t-1)}\right) \tag{20}$$

Here, MGU(·) represents the forward propagation of the multi-layer MGU network. We employ an optimized MGU cell architecture designed to enhance computational efficiency and model performance. This MGU cell selectively updates/resets hidden states via gating mechanisms to better capture long-term dependencies in sequential data.

To enable decoder focus on critical graph nodes at each step, we incorporate an attention mechanism. Specifically, a content-based attention mechanism processes: static features $E^{static}$, dynamic features $E^{dynamic}$ and MGU output $r^{(t)}$. The mechanism computes node relevance scores per decoding step to generate selection logits.

First, we transform dynamic features $E^{dynamic}$, static features $E^{static}$ and MGU output $r^{(t)}$ through linear projection layers, obtaining projected features $d^{(t)} \in R^{B \times D_h \times N}$, $e^{(t)} \in R^{B \times D_h \times N}$ and $q^{(t)} \in R^{B \times D_h}$. We then compute attention scores $u_i^{(t)}$, measuring the compatibility between decoder states and node:

$$u_i^{(t)} = v^T \tanh\left(e_i^{(t)} + q^{(t)} + d_i^{(t)}\right) \tag{21}$$

Where $v \in R^{D_h}$ is a learnable parameter vector, $e_i^{(t)}$ and $d_i^{(t)}$ denote the $i-th$ column vectors of $e^{(t)}$ and $d^{(t)}$ respectively, with $q^{(t)}$ broadcasted to match dimensions. The attention weights $\alpha^{(t)} \in R^{B \times N}$ are obtained by Softmax normalization of scores, indicating relative selection probabilities:

$$\alpha_i^{(t)} = \exp\left(u_i^{(t)}\right) \bigg/ \sum_{j \in N} \exp\left(u_j^{(t)}\right) \tag{22}$$

These weights generate context vectors aggregating node information weighted by decoder states.

To regulate the scale of logits and introduce nonlinearity, we apply activation to the attention scores and multiply by a learnable scaling factor, obtaining final node selection logits $l^{(t)} \in R^{B \times N}$:

$$l^{(t)} = C \cdot \tanh\left(u^{(t)}\right) \tag{23}$$

Where $u^{(t)} = \left[u_1^{(t)}, u_2^{(t)}, \ldots, u_N^{(t)}\right]^T$.

The decoder utilizes the attention-generated logits $l^{(t)}$ to compute node selection probability distribution through the Softmax function:

$$p^{(t)} = \text{Softmax}\left(l^{(t)}\right) \tag{24}$$

During the training phase, we employ a sampling-based action selection strategy. To ensure the validity of selected actions, we implement an action masking mechanism prior to generating probability distributions. The action mask dynamically filters out invalid action options (e.g., visited nodes or constraint-violating nodes) based on real-time environmental states. During model evaluation, we adopt a greedy action selection strategy.

To capture state evolution during decoding, the dynamic features $E^{dynamic}$ are updated at each decoding timestep.

### 4.3. Training strategy

As show in Algorithm 1, This approach employs an Asynchronous Advantage Actor-Critic framework enhanced with prioritized experience sampling and adaptive learning rate scheduling, specifically tailored for combinatorial optimization in vehicle routing problems. The core objective minimizes the expected completion time of coordinated truck-drone delivery routes, mathematically formulated as: $J(\theta) = E_{s \sim S}\left[E_{\pi \sim p_\theta(\cdot|s)} C(\pi/s)\right]$ where $\theta$ denotes policy parameters, $s$ represents problem instances sampled from distribution. and $\pi$ is the action sequence generated by the policy network.

To address computational challenges in large-scale routing optimization, we designed a hybrid training architecture decoupling environment interaction from parameter updates. As show in Algorithm 2 and Algorithm 3, The policy network (actor) and value network (critic) operate through parallel threads: the main thread generates trajectories via environment interaction, while auxiliary threads asynchronously update network parameters using gradients queued in a prioritized buffer. This design minimizes computational resource idling and accelerates convergence.

The policy gradient is derived using the REINFORCE algorithm with a critic-based baseline for variance reduction. For a batch of B problem instances $(\{s_i\}_{i=1}^{B})$, the gradient estimate is expressed as:

$$\nabla_\theta J(\theta) \approx \frac{1}{B}\sum_{i=1}^{B}\left(C(\pi_i/s_i)-b_\phi(s_i)\right)\nabla_\theta \log p_\theta(\pi_i/s_i) \tag{25}$$

where, $(b_\phi(s_i))$ denotes the critic network's predicted expected completion time, and $(\pi_i \sim p_\theta(\cdot|s_i))$ represents trajectories sampled from the policy. The critic network is trained by minimizing the mean squared error (MSE):

$$L(\phi) = \frac{1}{B}\sum_{i=1}^{B}\|b_\phi(s_i)-C(\pi_i|s_i)\|_2^2 \tag{26}$$

A dynamic prioritization mechanism filters high-impact experiences to accelerate the learning process. Let the advantage function be defined as:

$$(A(s_i,\pi_i) = C(\pi_i|s_i)-b_\phi(s_i)) \tag{27}$$

When the trajectory's average absolute advantage exceeds threshold $(\tau = 0.5)$, $\frac{1}{B}\sum_{i=1}^{B}|A(s_i,\pi_i)| > \tau$ prioritized immediate gradient updates are triggered. This prioritization focuses training on episodes where policy performance significantly deviates from critic estimates, efficiently allocating computational resources to challenging instances.

Network optimization employs the AdaBelief optimizer (Zhuang et al., 2020), which adaptively adjusts step sizes based on gradient magnitude and directional consistency.

$$m_t = \beta_1 m_{t-1} + (1-\beta_1)g_t, \tag{28}$$

$$v_t = \beta_2 v_{t-1} + (1-\beta_2)(g_t - m_t)^2 + , \tag{29}$$

$$\theta_{t+1} = \theta_t - \eta m_t/\sqrt{v_t} + \tag{30}$$

Where $g_t$ is the stochastic gradient, $\eta$ is the initial learning rate, and the hyperparameters are set to $\beta_1 = 0.9$, $\beta_2 = 0.999$, $= 10^{-16}$. Weight decay (0.01) and decoupled weight updates regularize the network.

The learning rate follows a cosine annealing schedule, completing five full cycles throughout the entire training period:

$$\eta_t = \eta_{min} + 1/2(\eta_{max} - \eta_{min})(1+\cos(t\pi/T_{max})) \tag{31}$$

Where $T_{max} = N_{train}/5$, $\eta_{min} = 0.01\eta_{max}$, $N_{train}$ is the total number of training cycles. This cyclical schedule promotes exploration by periodically resetting the learning rate, preventing premature convergence to suboptimal policies.

**Algorithm 1**: **Training Process**

Input:

    $\theta_0$ : Initial actor network parameters

    $\varphi_0$ : Initial critic network parameters

    $K$ : Maximum training epochs

    $\tau$ : Priority threshold

    $T$ : Maximum steps per episode (decode_len)

1  Initialize $actor(\pi_\theta)$ and critic $V_\varphi$ with

2  Start asynchronous $ActorUpdateThread(\theta_0)$ and $CriticUpdateThread(\varphi_0)$

3  for $epoch \leftarrow 1$ to $K$ do

4      $G \leftarrow GenerateTrainingData(n_{nodes})$

5      // Environment Interaction

6      $(S, avail) \leftarrow Env.Reset(G)$

7      $h \leftarrow InitializeMGU()$

8      $current_{time} \leftarrow 0$

9      for $t \leftarrow 0$ to $T-1$ do

10         $a_{tr}, logp_{tr}, h \leftarrow Actor_{\theta(S_{tr}, h, avail_{tr})}$

11         Update $avail_{dr}$ based on $a_{tr}$

12         $a_{dr}, logp_{dr}, h \leftarrow Actor_{\theta(S_{dr}, h, avail_{dr})}$

13         $time_{step} \leftarrow \min(t_{tr}, t_{dr}, remaining_{time})$

14         $current_{time} += time_{step}$

15         $S, avail \leftarrow S', avail'$

16      end for

17      $A \leftarrow R - V_{\varphi(static, w)}$

18      if $mean(|A|) > \tau$ then

19         $actor_{loss} \leftarrow -mean\left[\sum_{t=0}^{T-1} A \cdot \left(\log \pi(a_{tr}^t) + \log \pi(a_{dr}^t)\right)\right]$

20         $critic_{loss} \leftarrow mean(A^2)$

21         Send $actor_{loss}$ to $ActorQueue$

22         Send $critic_{loss}$ to $CriticQueue$

23      end if

24      if $epoch \bmod S_{interval} = 0$ then

| 25 | $R_{val} \leftarrow Validate(\theta, \varphi)$ |
| --- | --- |
| 26 |     if $R_{val} < R_{best}$ then |
| 27 |         $R_{best} \leftarrow R_{val}$ |
| 28 |         $\theta^* \leftarrow \theta$ |
| 29 |         $\varphi^* \leftarrow \varphi$ |
| 30 |         $SaveModel(\theta^*, \varphi^*)$ |
| 31 |     end if |
| 32 | end if |
| 33 | end for |

**Algorithm 2: ActorUpdateThread**

| 1 | while training not terminated do |
| --- | --- |
| 2 |     if $ActorQueue$ not empty then |
| 3 |         $actor_{loss} \leftarrow ActorQueue.get(\ )$ |
| 4 |         $\theta \leftarrow AdaBeliefUpdate(\theta, actor_{loss})$ |
| 5 |         $UpdateActorLearningRate(\theta)$ |
| 6 |     end if |
| 7 | end while |

**Algorithm 3: CriticUpdateThread**

| 1 | while training not terminated do |
| --- | --- |
| 2 |     if $CriticQueue$ not empty then |
| 3 |         $critic_{loss} \leftarrow CriticQueue.get(\ )$ |
| 4 |         $\varphi \leftarrow AdaBeliefUpdate(\varphi, critic_{loss})$ |
| 5 |         $UpdateCriticLearningRate(\varphi)$ |
| 6 |     end if |
| 7 | end while |

# 5 Computational Experiments

To comprehensively evaluate the performance of this deep reinforcement learning method, this chapter compares our algorithm with traditional heuristic algorithms and high-performance reinforcement learning algorithms in the field. In all experiments, we assume the drone is twice as fast as the truck (α = 2) and has unlimited flight range.

Few standard datasets exist for TSP-D. (Bogyrbayeva et al., 2023) developed a new dataset based on (Haider et al., 2019) 's study , which has subsequently been adopted in TSP-D research (Mahmoudinazlou and Kwon, 2024), This collection contains two TSP-D instance subsets ("Random" and "Amsterdam") with unlimited drone range, differing in instance generation methodologies. In the first instance subset, uniform distributions over [0, 1] $\times$ [0, 1] and [0, 100] $\times$ [0, 100] are employed to sample x and y coordinates of depots and customer nodes, respectively. This configuration ensures depot positioning at the lower-left corner. The generation method resembles that proposed by (Agatz et al., 2018) We present 100 instance samples for node sizes of 20, 50, and 100. The second instance subset comprises 100 samples with node sizes of 10, 20 and 50. In this subset, depots are randomly selected from the nodes. Additionally, we evaluated another uniform dataset for TSP-D maintained in the TSBlib project on GitHub.

We employ the following formula to report relative gaps between solutions:

$$\text{Gap}_i = \left(Z_i - Z_i^*\right) / Z_i^* \times 100\% \tag{32}$$

Here, $Z_i$ denotes the cost of a specific solution for instance $i$, and $Z_i^*$ represents the solution method widely recognized as high-quality for instance $i$ among existing comparative approaches. We then report $Gap_i$ as the average of all $Gap$ values under identical configurations.

5.1. Model Inference Performance

For reinforcement learning algorithms, since this study builds upon AM (Kool et al., 2018) and HM (Bogyrbayeva et al., 2023), natural comparisons should be made with them. Notably, AM has demonstrated inferior performance to HM in both solution accuracy and speed across varying node scales (Bogyrbayeva et al., 2023), thus excluded from our comparison. Inference was conducted using a single RTX 4090 (24GB) GPU with 12 vCPU Intel Xeon Platinum 8352V CPU @ 2.10GHz, under PyTorch 1.10.0, Python 3.8 (Ubuntu 20.04), and CUDA 11.3.

For heuristic algorithms, (Bogyrbayeva et al., 2023) extended the "TSP-ep-all" algorithm by (Agatz et al., 2018) to DPS by adopting the "divide-and-conquer heuristic" (DCH) from (Poikonen et al., 2019) . We compared the performance of "TSP-ep-all" and DPS. (Mahmoudinazlou and Kwon, 2024) proposed a Hybrid Genetic Algorithm with Type-Aware Chromosomes that integrates local search and dynamic programming to solve TSP-D, showing strong performance on "Amsterdam" and "Random" datasets, which we also compared against. The key innovation in their approach lies in observing that escape strategies improve solution quality while increasing runtime. This demonstrates effective mitigation of metaheuristic methods' inherent local optima issues. Consequently, we only compared HGA-TAC$^+$ implementations employing escape strategies. All three heuristic algorithms were implemented in Julia and evaluated using an AMD EPYC 9654 96-core processor.

For the greedy policy in reinforcement learning, instances are not evaluated in batches but sequentially processed one-by-one on the GPU. Heuristic methods follow the same strategy but utilize a single CPU thread instead of GPU. For batch sampling strategies, HM (n_samples) (where n_samples denotes batch size) demonstrates that increasing n_samples generally improves solution accuracy while reducing computational speed. Each algorithm was executed 10 times, with the best reward and computation time per instance averaged across trials.

We initially evaluated using the "Random" dataset. To rigorously evaluate the performance of our proposed model for the TSP-D, we executed each algorithm 10 times and calculated the average of the cost and computation times for each instance. we conducted extensive computational experiments on benchmark instances with varying network sizes (N=11, 20, 50, and 100 nodes). Performance was assessed based on two primary metrics: the total solution cost (lower values indicate better performance) and the computational time required (in seconds, lower values indicate greater efficiency). We employed TSP-ep-all, an approach known for generating high-quality solutions, as the baseline for comparison. The percentage gap (Gap) relative to this baseline is reported to quantify the deviation in solution quality.

We first examine the performance of rapid decision-making strategies, including DPS/10, HM (greedy), and our own greedy implementation (Ours(greedy)). As show in Table 1, These methods prioritize computational speed, consistently delivering solutions within fractions of a second across all instance sizes (e.g., $\leq$ 0.01s for greedy methods, $\leq$ 0.27s for DPS/10 at N=100). However, this velocity is achieved at the expense of solution quality. Both HM (greedy) and Ours (greedy) exhibit similar performance, yielding positive Gaps ranging from approximately 0.7% to 5.4% compared to the baseline. DPS/10 shows comparable Gaps (3.70% to 5.90%). While extremely fast, these methods demonstrate the limitations of purely greedy strategies in navigating the complex TSP-D solution landscape.

The core of our evaluation lies in comparing Ours (sampling_k) against the HM (sampling_k) using equivalent number of samples for batch sampling (k=100, 1200, 2400, 4800). For any given number of samples and problem size, ours (sampling_k) consistently achieves lower costs than HM (sampling_k). This advantage is particularly evident in larger instances. For example, at N=50 and

k=4800, our method yields a cost of 391.51 (Gap: -1.42%), significantly outperforming HM's 396.31 (Gap: -0.22%). Similarly, at N=100 and k=4800, our method achieves a 1.22% Gap (Cost: 542.21) compared to HM's 1.63% Gap (Cost: 544.42). This suggests that our RL framework integrates the sampled information more effectively to guide the policy towards higher-quality solutions.

Concurrent with superior solution quality, our algorithm exhibits enhanced computational efficiency. When comparing Ours (sampling_k) and HM (sampling_k) at identical N and k, our method consistently requires less computational time. For instance, solving the N=100 instance with k=4800 takes 8.43s using our method, whereas HM requires 9.54s, representing an approximate 11.6% reduction in runtime.

Increasing the sampling width (k) predictably improves solution quality for RL method, when compared against the strong heuristic HGVAC$^+$, our sampling method presents a compelling alternative. While HGVAC$^+$ performs well, particularly on smaller instances (N=11, 20) where it achieves negative Gaps relatively quickly, our method (Ours (sampling_k) with sufficient k) often matches or surpasses its solution quality on larger instances while offering significant computational savings. For N=50, Ours(sampling_4800) achieves a better Gap (-1.42%) in nearly half the time (2.41s vs 4.65s) compared to HGVAC$^+$ (Gap 0.39%). For N=100, Ours (sampling_4800) delivers a marginally better Gap (1.22% vs 1.53%) while being substantially faster (8.43s vs 14.83s). This suggests our method scales more effectively in terms of maintaining high-quality solutions within reasonable time limits as problem complexity increases.

**Table 1**

TSP-D results on "Random" locations dataset

|  | N=11 | | | N=20 | | | N=50 | | | N=100 | | |
| --- | --- | --- | --- | --- | --- | --- | --- | --- | --- | --- | --- | --- |
| Method | cost | Gap | Time(s) | cost | Gap | Time(s) | cost | Gap | Time(s) | cost | Gap | Time(s) |
| TSP-ep-all(Baseline) | 230.07 | 0.00% | 0.01 | 281.62 | 0.00% | 0.11 | 397.17 | 0.00% | 23.95 | 535.67 | 0.00% | 2352.53 |
| DPS/10 |  |  |  | 292.05 | 3.70% | 0.02 | 420.61 | 5.90% | 0.05 | 565.14 | 5.50% | 0.27 |
| HGVAC$^+$ | 227.45 | -1.14% | 0.52 | 279.88 | -0.62% | 1.22 | 398.72 | 0.39% | 4.65 | 543.88 | 1.53% | 14.83 |
| HM(greedy) | 233.21 | 1.36% | 0.00 | 285.54 | 1.39% | 0.00 | 408.84 | 2.94% | 0.01 | 564.42 | 5.37% | 0.01 |
| Ours(greedy) | 231.67 | 0.69% | 0.00 | 285.80 | 1.48% | 0.00 | 407.04 | 2.49% | 0.00 | 564.36 | 5.36% | 0.01 |
| HM(sampling_100) | 230.10 | 0.01% | 0.11 | 282.93 | 0.46% | 0.13 | 399.59 | 0.61% | 0.35 | 550.13 | 2.70% | 0.76 |
| Ours(sampling_100) | 229.05 | -0.45% | 0.10 | 282.10 | 0.17% | 0.13 | 396.41 | -0.19% | 0.35 | 550.89 | 2.84% | 0.74 |
| HM(sampling_1200) | 229.22 | -0.37% | 0.14 | 282.13 | 0.18% | 0.18 | 397.38 | 0.05% | 0.70 | 546.01 | 1.93% | 2.57 |
| Ours(sampling_1200) | **228.55** | -0.66% | 0.13 | **280.80** | -0.29% | 0.17 | **392.94** | -1.06% | 0.68 | **544.41** | 1.63% | 2.21 |
| HM(sampling_2400) | 229.12 | -0.42% | 0.18 | 281.84 | 0.08% | 0.28 | 397.01 | -0.04% | 1.41 | 545.13 | 1.77% | 4.90 |
| Ours(sampling_2400) | **228.36** | -0.74% | 0.15 | **280.67** | -0.34% | 0.23 | **392.07** | -1.28% | 1.27 | **543.18** | 1.40% | 4.22 |
| HM(sampling_4800) | 228.93 | -0.50% | 0.29 | 281.67 | 0.02% | 0.50 | 396.31 | -0.22% | 2.59 | 544.42 | 1.63% | 9.54 |
| Ours(sampling_4800) | **228.30** | -0.77% | 0.21 | **280.44** | -0.42% | 0.42 | **391.51** | -1.42% | 2.41 | **542.21** | 1.22% | 8.43 |

As show in Table 2. On the uniform dataset, our proposed reinforcement learning algorithm demonstrates superior performance compared to the HM method across TSP-D instances of varying scales (N=10, 20, 50). Utilizing the sampling_4800, our approach consistently yielded lower costs, achieving negative Gaps of -0.12%, -1.10%, and -0.58% for N=10, N=20, and N=50, respectively, relative to the computationally intensive HM (4800) baseline. This signifies a notable improvement in solution quality. Crucially, this enhancement was realized with comparable or marginally reduced computational times (e.g., 2.47s vs. 2.57s for N=50). While both algorithms improve solution quality with increased sampling, our method exhibits greater effectiveness, achieving superior solutions faster. Notably, for N=20 and N=50, our algorithm with only 1200 or 2400 samples already surpasses the cost performance of the HM (4800) baseline (indicated by negative Gaps: -0.89%/-1.10% for N=20, -0.31%/-0.50% for N=50). **Fig. 4** presents detailed routing visualizations on the 20-node uniform dataset, clearly demonstrating that our model with sampling strategy outperforms HM in nearly every instance.

**Table 2**

TSP-D results on "uniform" locations dataset

| Method | N=11 | | | N=20 | | | N=50 | | |
| --- | --- | --- | --- | --- | --- | --- | --- | --- | --- |
| | cost | Gap | Time | cost | Gap | Time | cost | Gap | Time |
| HM(greedy) | 228.38 | 0.28% | 0.00 | 277.87 | 0.66% | 0.00 | 426.93 | 4.26% | 0.01 |
| HM(sampling_100) | 228.38 | 0.28% | 0.09 | 276.95 | 0.33% | 0.12 | 413.22 | 0.91% | 0.35 |
| HM(sampling_1200) | 227.98 | 0.10% | 0.15 | 276.09 | 0.02% | 0.17 | 410.57 | 0.27% | 0.69 |
| HM(sampling_2400) | 227.75 | 0.00% | 0.18 | 276.09 | 0.02% | 0.26 | 409.94 | 0.11% | 1.31 |
| HM(sampling_4800)(Baseline) | 227.75 | 0.00% | 0.29 | 276.05 | 0.00% | 0.50 | 409.48 | 0.00% | 2.57 |
| Ours(greedy) | 228.38 | 0.28% | 0.00 | 279.23 | 1.15% | 0.00 | 425.68 | 3.96% | 0.01 |
| Ours(sampling_100) | 227.76 | 0.00% | 0.11 | 275.51 | -0.19% | 0.14 | 412.42 | 0.72% | 0.35 |
| Ours(sampling_1200) | 227.49 | -0.12% | 0.14 | 273.60 | -0.89% | 0.19 | 408.23 | -0.31% | 0.74 |
| Ours(sampling_2400) | 227.49 | -0.12% | 0.18 | 273.00 | -1.10% | 0.24 | 407.43 | -0.50% | 1.28 |
| Ours(sampling_4800) | 227.49 | -0.12% | 0.29 | 273.00 | -1.10% | 0.46 | 407.09 | -0.58% | 2.47 |

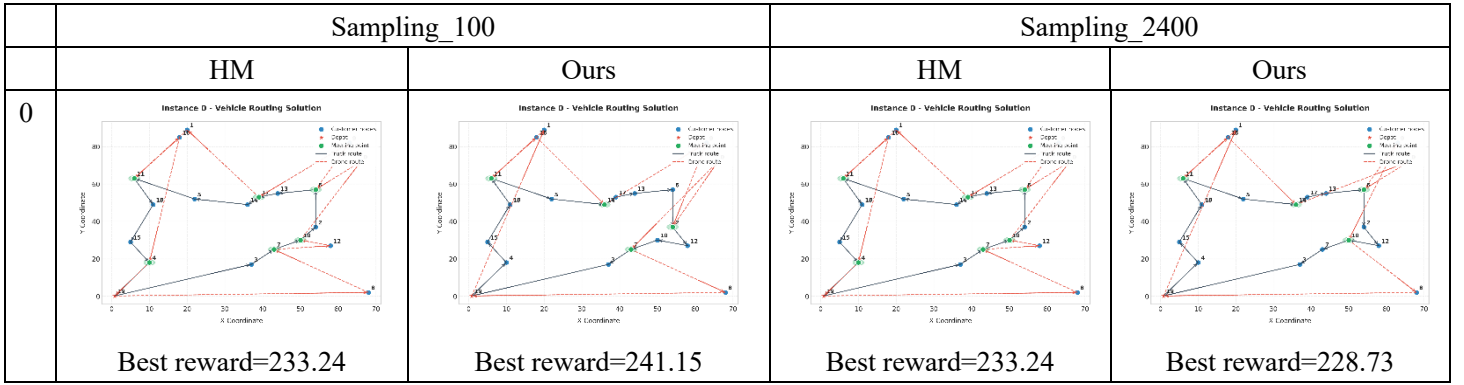

| | Sampling_100 | | Sampling_2400 | |
| --- | --- | --- | --- | --- |
| | HM | Ours | HM | Ours |
| 0 | Best reward=233.24 | Best reward=241.15 | Best reward=233.24 | Best reward=228.73 |

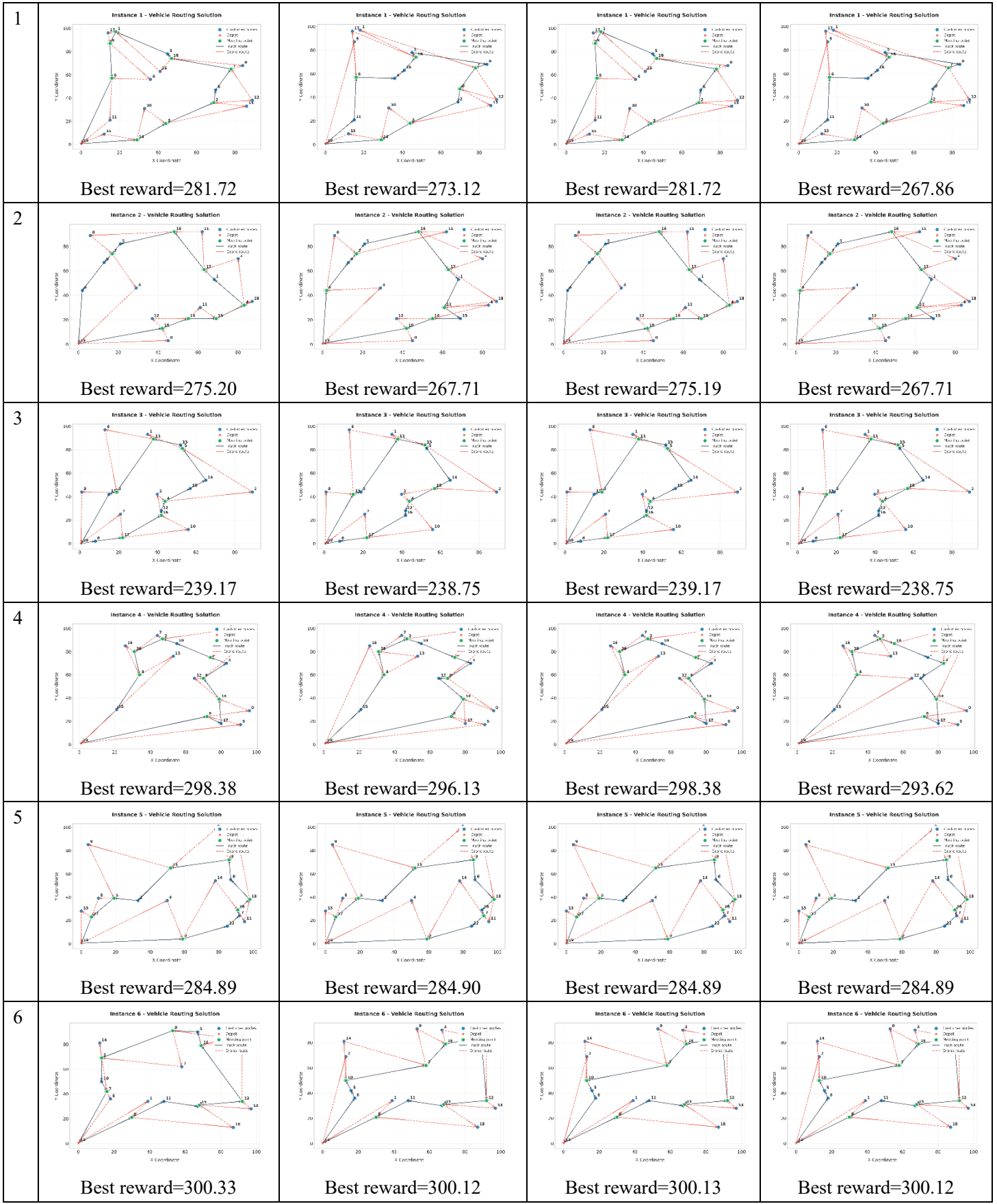

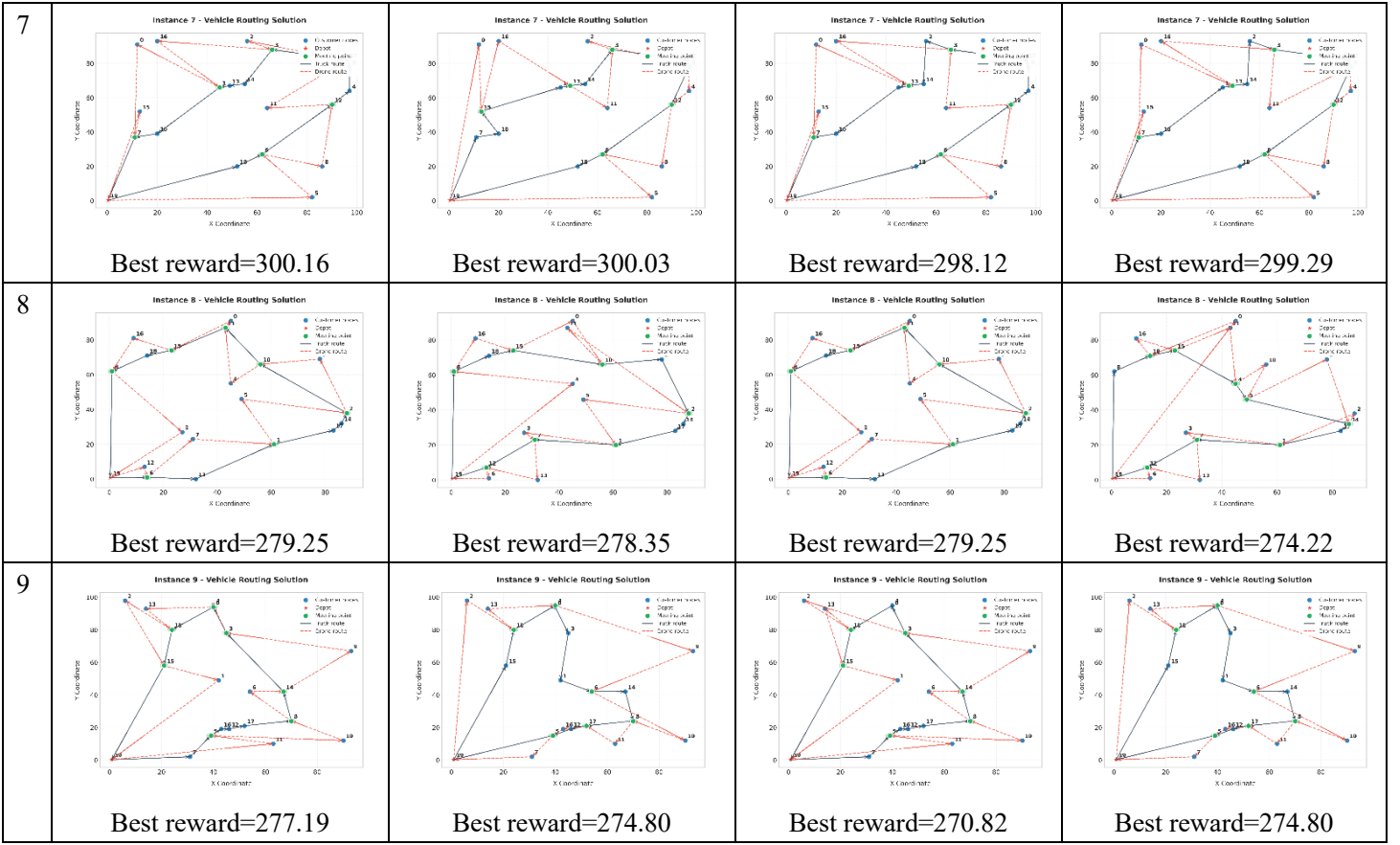

**Fig. 4.** Test results of our model and HM on the "uniform" dataset position with 20 nodes

We also compared the model performance on the Amsterdam dataset, As show in Table 3, since the Amsterdam dataset has 10 nodes, the performance differences of multiple models on 11 nodes are not compared. For the N=20 instances, our proposed algorithm (Ours(sampling_4800)) demonstrates highly competitive performance. It achieves a solution cost of 2.34, marginally outperforming the TSP-ep-all baseline (Gap: -0.70%) and matching the heuristic HGVAC+. Notably, its computation time (0.48s) is significantly faster than HGVAC$^+$ (1.81s) and comparable to the HM reinforcement learning method (0.50s), while being only moderately slower than the baseline's 0.12s. Although DPS methods offer extreme speed (≤0.05s), their solution quality is substantially compromised (Gap: 33.20%), highlighting a clear trade-off. Our method effectively balances near-optimal solution quality with efficient computation time at this scale.

This advantage in computational efficiency becomes more pronounced for the larger N=50 instances. Our algorithm yields a cost of 3.29, representing the best performance among the RL methods (Gap: 0.76%) and remaining highly competitive with the baseline cost (3.27). Crucially, our execution time (2.47s) remains efficient and similar to HM (2.58s), offering a nearly 10-fold speed advantage over the baseline (24.53s) and a 2.5-fold advantage over HGVAC$^+$ (6.20s). These results underscore our method's strong scalability, delivering high-quality solutions with significantly reduced computational burden compared to both the heuristic baseline and other competitive approaches as problem size increases.

**Table 3**

TSP-D results on "Amsterdam" locations dataset

|  | N=20 | | | N=50 | | |
| --- | --- | --- | --- | --- | --- | --- |
| Method | cost | Gap | Time | cost | Gap | Time |
| TSP-ep-all(Baseline) | 2.36 | 0.00% | 0.12 | 3.27 | 0.00% | 24.53 |
| DPS/10 | 3.14 | 33.20% | 0.05 | 3.80 | 16.49% | 0.08 |
| DPS/25 |  |  | 0.00 | 4.23 | 29.46% | 0.07 |
| HGVAC$^+$ | 2.34 | -0.79% | 1.81 | 3.33 | 2.10% | 6.20 |
| HM(sampling_4800) | 2.38 | 1.00% | 0.50 | 3.31 | 1.37% | 2.58 |
| Ours(sampling_4800) | 2.34 | -0.70% | 0.48 | 3.29 | 0.76% | 2.47 |

## 5.2. Model Training Performance

We found that compared to the improvement in inference speed, the model's improvement in training speed and convergence speed is more significant. Specifically, our model can achieve better results within the same number of epochs, and the training speed per epoch is also significantly increased. We compared the differences in training speed and convergence speed with HM on a single 4090 GPU.

To evaluate the training characteristics of the proposed model (Ours) and the baseline model HM in solving TSP-D problems of different scales, we conducted detailed comparative experiments on instances with 11, 20, 50 and 100 nodes respectively. As show in Fig. 5 and Fig. 6, both models were trained for 10,000 epochs, and key performance indicators, including the reward value (Reward, representing path cost here, lower is better) and training time, were recorded every 200 epochs. Analyzing the training curves (reward value change over epochs), both models exhibit the typical characteristics of reinforcement learning, i.e., the reward value shows an overall downward trend as training progresses, indicating continuous optimization of the model's policy. However, our model (Ours) demonstrates significant advantages in terms of convergence speed and final solution quality. Taking n=11 as an example, the Ours model obtained a much lower reward value than the HM model in the initial stage (Epoch 0) (311.7 vs 409.5), and reached a reward value close to the final level (~253) around 1200 epochs, while the HM model's reward value was still above 260 during the same period and required more epochs to stabilize. This advantage of rapid convergence is even more pronounced on larger-scale problems (n=20, 50, 100). For example, at n=100, the Ours model could rapidly reduce the reward value from about 958 to around 700 in the early training stages (around 1000-1400 epochs), while the HM model required a longer exploration period to reach a similar level. At the end of training (Epoch 10000), for all tested node scales, the final average reward value obtained by the Ours model was lower than that of the HM model (n=11:

242.7 vs 245.3; n=20: 304.4 vs 312.5; n=50: 477.5 vs 487.8; n=100: 677.5 vs 688.0), which demonstrates the effectiveness of our model in finding better TSP-D solutions. Furthermore, our proposed model exhibits training stability significantly superior to the HM baseline model. Across all tested node scales (n=11, 20, 50, 100), the evolution of the Ours model's reward value over training epochs shows smaller fluctuations, especially in the middle and later stages of training, its learning curve is smoother, and rarely exhibits the drastic performance oscillations or significant temporary policy degradation phenomena observed in the HM model.

In terms of computational efficiency, our model (Ours) also shows significant superiority. By comparing the recorded average training time per epoch, it can be found that under all node scales, the time consumption per epoch of the Ours model is significantly lower than that of the HM model. For example, for n=11, the average epoch time of the Ours model is about 0.13 seconds, while the HM model is about 0.27-0.30 seconds. When the scale increases to n=100, the epoch time of the Ours model is about 0.9-1.1 seconds, while the HM model requires 1.3-1.7 seconds. The cumulative effect of this improvement in per-epoch efficiency is significant, leading to the total cumulative time of the Ours model being much less than the HM model when completing the same number of training epochs (10,000 Epochs). Specifically, for n=11, Ours total time consumption is about 1438 seconds, HM time consumption is about 2838 seconds. For n=100, Ours total time consumption is about 10241 seconds (about 2.8 hours), while HM time consumption is 16730 seconds (about 4.6 hours). This indicates that the Ours model not only trains faster but also has lower computational costs. From the perspective of scalability, although the per-epoch time consumption and total time consumption of both models inevitably increase with the increase in problem size, the Ours model always maintains a relative time efficiency advantage. Overall, our model can not only converge faster to higher-quality solutions during the training process but also has significant advantages in computational resource consumption, especially when dealing with larger-scale problems, this dual improvement in efficiency and effectiveness makes it a more potential reinforcement learning solution for solving TSP-D problems.

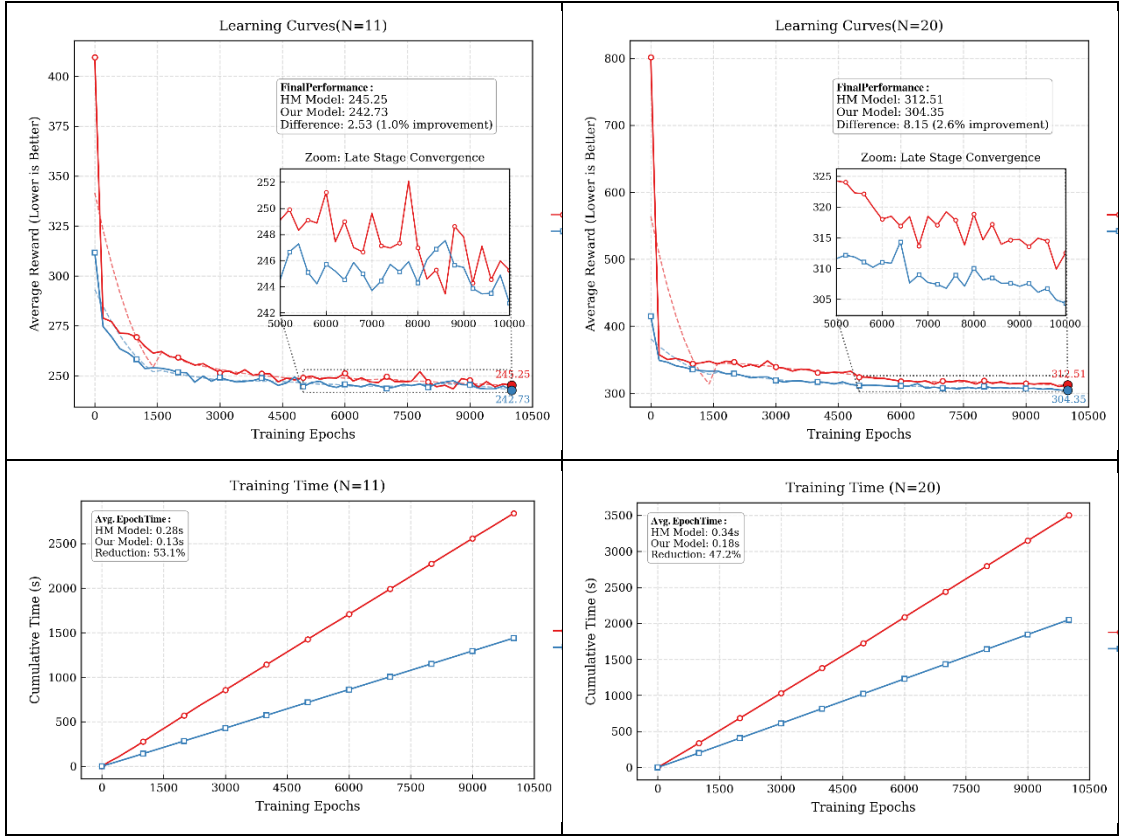

**Fig. 5.** The training learning curves and training time graphs of our model and the HM model at 11, 20 nodes

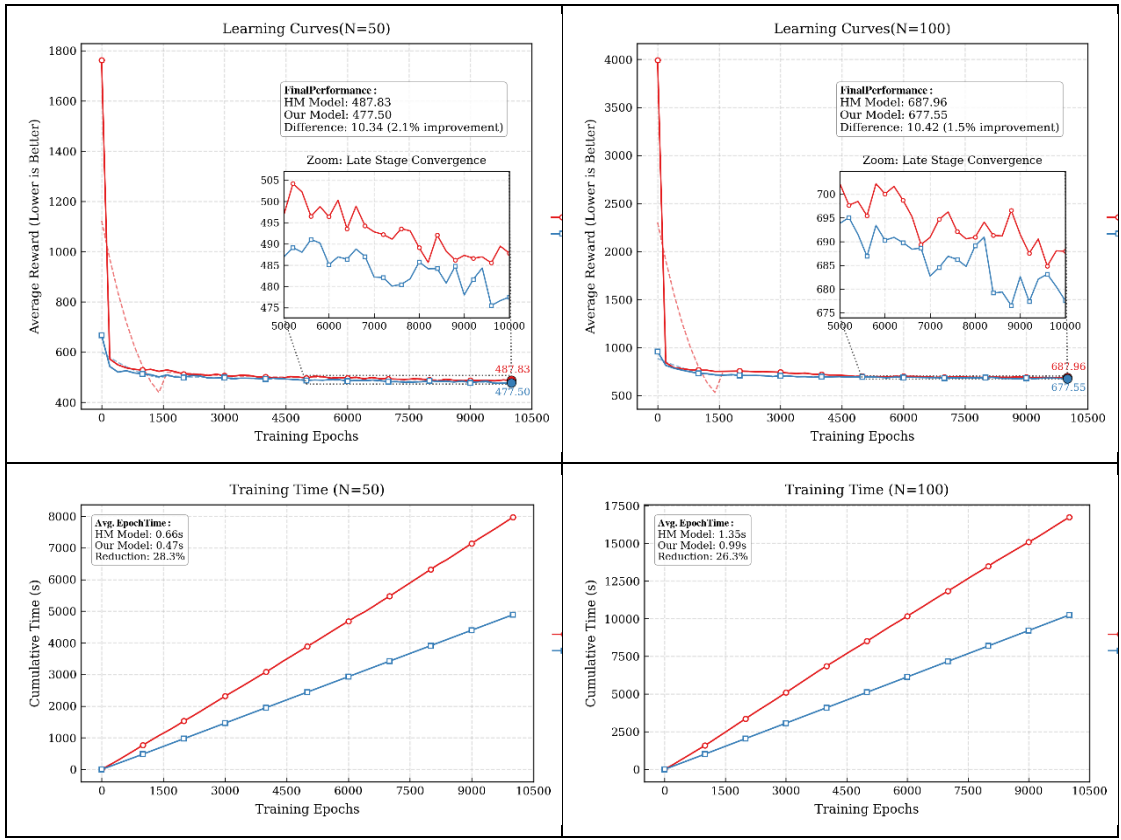

**Fig. 6.** The training learning curves and training time graphs of our model and the HM model at 50, 100 nodes

# 6 Conclusions

In this study, a novel end-to-end deep reinforcement learning framework was proposed to solve TSP-D. The contributions of this study are as follows: First, a Transformer-inspired encoder architecture with optimized k-nearest neighbors sparse attention was developed to efficiently extract crucial structural patterns within TSP-D instances. Second, we introduced dynamic expander graph Masking and global node features to capture the intricate spatial and collaborative dynamics of the problem. Third, an efficient MGU based decoder was implemented alongside an advanced asynchronous advantage actor-critic training strategy, incorporating prioritized experience replay and adaptive learning rate adjustments to enhance training efficiency.

We evaluated the performance of the proposed model in solving the TSP-D problem using three sets of benchmark instances. Compared with state-of-the-art heuristic algorithms (TSP-ep-all, DPS, HGVAC$^+$) and the reinforcement learning algorithm (HM) in the field. Experimental results demonstrate that our proposed method generates high-quality solutions across various TSP-D instances, exhibiting competitive solution quality compared to leading heuristic algorithms while demonstrating superior generalization capabilities on large-scale problems. Notably, when handling medium to large scale instances with 20, 50 and 100 nodes, our method achieves high-quality solutions across nearly all datasets while maintaining low computational costs. Furthermore, it significantly reduces training time compared to state-of-the-art reinforcement learning algorithms during model training. The proposed method offers valuable insights for combinatorial optimization problems in multi-agent collaborative decision-making. It also introduces novel ideas and technical support for developing intelligent decision-making systems in logistics.

This study had some limitations. While demonstrating strong performance up to N=100, the framework's scalability and performance on significantly larger or more densely connected instances warrant further investigation. Moreover, the current model focuses primarily on core routing optimization without incorporating more nuanced real-world operational constraints. Therefore, in future work, we will focus on enhancing the scalability and robustness of the proposed architecture, possibly through exploring more advanced graph representation techniques or hierarchical learning approaches. Additionally, validating the model's performance on diverse real-world datasets, potentially incorporating operational constraints such as varying drone battery endurance, payload capacities, or specific customer time windows, will be critical for bridging the gap towards practical deployment in logistics systems. Extending this framework to tackle other complex multi-agent or heterogeneous fleet routing problems also represents a valuable direction for future research.

## Acknowledgments

This work was supported by the National Natural Science Foundation of China (Nos: 52105507 and 52275476) and Specialized Research Fund for Chongqing Technology Innovation and Application Development (No: CSTB2022TIAD-KPX0061).

## Declaration of Interest statement

Authors declare that there is no conflict of interest due to the publication of this paper.

# References

Agatz, N., Bouman, P., Schmidt, M., 2018. Optimization Approaches for the Traveling Salesman Problem with Drone. *Transportation Science* 52(4), 965-981.

Bello, I., Pham, H., Le, Q.V., Norouzi, M., Bengio, S., 2016. Neural combinatorial optimization with reinforcement learning. *arXiv preprint arXiv:1611.09940*.

Bengio, Y., Lodi, A., Prouvost, A., 2021. Machine learning for combinatorial optimization: A methodological tour d'horizon. *European Journal of Operational Research* 290(2), 405-421.

Bogyrbayeva, A., Yoon, T., Ko, H.B., Lim, S., Yun, H.Y.K., Kwon, C., 2023. A deep reinforcement learning approach for solving the Traveling Salesman Problem with Drone. *Transp. Res. Pt. C-Emerg. Technol.* 148, 19.

Cho, K., Van Merriënboer, B., Gulcehre, C., Bahdanau, D., Bougares, F., Schwenk, H., Bengio, Y., 2014. Learning phrase representations using RNN encoder-decoder for statistical machine translation. *arXiv preprint arXiv:1406.1078*.

Costa, L., Contardo, C., Desaulniers, G., 2019. Exact Branch-Price-and-Cut Algorithms for Vehicle Routing. *Transportation Science* 53(4), 946-985.

Deudon, M., Cournut, P., Lacoste, A., Adulyasak, Y., Rousseau, L.M., 2018. Learning Heuristics for the TSP by Policy Gradient, *15th International Conference on the Integration of Constraint Programming, Artificial Intelligence, and Operations Research (CPAIOR)*. Springer International Publishing Ag, Delft, NETHERLANDS, pp. 170-181.

Duan, J., Luo, H., Wang, G., 2025. Approaches to the truck-drone routing problem: A systematic review. *Swarm and Evolutionary Computation* 92, 101825.

Gebreyesus, G., Fellek, G., Farid, A., Fujimura, S., Yoshie, O., 2023. Gated-Attention Model with Reinforcement Learning for Solving Dynamic Job Shop Scheduling Problem. *IEEJ Trans. Electr. Electron. Eng.* 18(6), 932-944.

Gu, R.X., Poon, M., Luo, Z.H., Liu, Y., Liu, Z., 2022. A hierarchical solution evaluation method and a hybrid algorithm for the vehicle routing problem with drones and multiple visits. *Transp. Res. Pt. C-Emerg. Technol.* 141, 23.

Haider, Z., Charkhgard, H., Kim, S.W., Kwon, C., 2019. Optimizing the relocation operations of free-floating electric vehicle sharing systems. *Available at SSRN 3480725*.

Ham, A.M., 2018. Integrated scheduling of $m$-truck, $m$-drone, and $m$-depot constrained by time-window, drop-pickup, and $m$-visit using constraint programming. *Transp. Res. Pt. C-Emerg. Technol.* 91, 1-14.

Hopfield, J.J., Tank, D.W., 1985. "Neural" computation of decisions in optimization problems. *Biological Cybernetics* 52(3), 141-152.

Joshi, C.K., Laurent, T., Bresson, X., 2019. An efficient graph convolutional network technique for the travelling salesman problem. *arXiv preprint arXiv:1906.01227*.

Karimi-Mamaghan, M., Mohammadi, M., Meyer, P., Karimi-Mamaghan, A.M., Talbi, E., 2022. Machine learning at the service of meta-heuristics for solving combinatorial optimization problems: A state-of-the-art. *European Journal of Operational Research* 296(2), 393-422.

Kitjacharoenchai, P., Ventresca, M., Moshref-Javadi, M., Lee, S., Tanchoco, J.M.A., Brunese, P.A., 2019. Multiple traveling salesman problem with drones: Mathematical model and heuristic approach. *Computers & Industrial Engineering* 129, 14-30.

Kong, F.H., Jiang, B., 2024. Delivery optimization for collaborative truck-drone routing problem considering vehicle obstacle avoidance. *Computers & Industrial Engineering* 198, 14.

Kool, W., Van Hoof, H., Welling, M., 2018. Attention, learn to solve routing problems! *arXiv preprint arXiv:1803.08475*.

Kuo, R.J., Lu, S.H., Lai, P.Y., Mara, S.T.W., 2022. Vehicle routing problem with drones considering time windows. *Expert Systems with Applications* 191, 19.

Laporte, G., 1992. The traveling salesman problem: An overview of exact and approximate algorithms. *European Journal of Operational Research* 59(2), 231-247.

Li, K., Zhang, T., Wang, R., 2021. Deep Reinforcement Learning for Multiobjective Optimization. *IEEE Transactions on Cybernetics* 51(6), 3103-3114.

Li, Y., Liu, M., Jiang, D.D., 2022. Application of Unmanned Aerial Vehicles in Logistics: A Literature Review. *Sustainability* 14(21), 18.

Liu, M., Wang, Z., Li, J., 2022. A deep reinforcement learning algorithm for large-scale vehicle routing problems, *International Conference on Electronic Information Technology (EIT 2022)*. SPIE, pp.

824-829.

Lu, H., Zhang, X., Yang, S., 2019. A learning-based iterative method for solving vehicle routing problems, *International conference on learning representations*.

Lu, J., Liu, Y.M., Jiang, C.M., Wu, W.W., 2025. Truck-drone joint delivery network for rural area: Optimization and implications. *Transp. Policy* 163, 273-284.

Ma, Q., Ge, S., He, D., Thaker, D., Drori, I., 2019. Combinatorial optimization by graph pointer networks and hierarchical reinforcement learning. *arXiv preprint arXiv:1911.04936*.

Mahmoudinazlou, S., Kwon, C., 2024. A hybrid genetic algorithm with type-aware chromosomes for Traveling Salesman Problems with Drone. *European Journal of Operational Research* 318(3), 719-739.

Montaña, L.C., Malagon-Alvarado, L., Miranda, P.A., Arboleda, M.M., Solano-Charris, E.L., Vega-Mejía, C.A., 2021. A novel mathematical approach for the Truck-and-Drone Location-Routing Problem, *3rd International Conference on Industry 4.0 and Smart Manufacturing (ISM)*. Elsevier Science Bv, Upper Austria Univ Appl Sci, Hagenberg Campus, Linz, AUSTRIA, pp. 1378-1391.

Murray, C.C., Chu, A.G., 2015. The flying sidekick traveling salesman problem: Optimization of drone-assisted parcel delivery. *Transp. Res. Pt. C-Emerg. Technol.* 54, 86-109.

Nazari, M., Oroojlooy, A., Snyder, L., Takác, M., 2018. Reinforcement learning for solving the vehicle routing problem. *Advances in neural information processing systems* 31.

Nguyen, M.A., Dang, G.T.H., Hà, M.H., Pham, M.T., 2022. The min-cost parallel drone scheduling vehicle routing problem. *European Journal of Operational Research* 299(3), 910-930.

Peng, B., Wang, J., Zhang, Z., 2019. A deep reinforcement learning algorithm using dynamic attention model for vehicle routing problems, *International Symposium on Intelligence Computation and Applications*. Springer, pp. 636-650.

Poikonen, S., Golden, B., Wasil, E.A., 2019. A Branch-and-Bound Approach to the Traveling Salesman Problem with a Drone. *INFORMS J. Comput.* 31(2), 335-346.

Poikonen, S., Wang, X.Y., Golden, B., 2017. The vehicle routing problem with drones: Extended models and connections. *Networks* 70(1), 34-43.

Pourmohammadreza, N., Jokar, M.R.A., Van Woensel, T., 2025. Last-mile logistics with alternative delivery locations: A systematic literature review. *Results in Engineering* 25, 104085.

Rave, A., Fontaine, P., Kuhn, H., 2023. Drone location and vehicle fleet planning with trucks and aerial drones. *European Journal of Operational Research* 308(1), 113-130.

Sacramento, D., Pisinger, D., Ropke, S., 2019. An adaptive large neighborhood search metaheuristic for the vehicle routing problem with drones. *Transp. Res. Pt. C-Emerg. Technol.* 102, 289-315.

Schermer, D., Moeini, M., Wendt, O., 2018. Algorithms for Solving the Vehicle Routing Problem with Drones, In: Nguyen, N.T., Hoang, D.H., Hong, T.-P., Pham, H., Trawiński, B. (Eds.), *Intelligent Information and Database Systems*. Springer International Publishing, Cham, pp. 352-361.

Shi, C., 2022. Pointer Network Solution Pool : Combining Pointer Networks and Heuristics to Solve TSP Problems, *2022 3rd International Conference on Computer Vision, Image and Deep Learning & International Conference on Computer Engineering and Applications (CVIDL & ICCEA)*, pp. 1236-1242.

Shirzad, H., Velingker, A., Venkatachalam, B., Sutherland, D.J., Sinop, A.K., 2023. EXPHORMER: Sparse Transformers for Graphs, *40th International Conference on Machine Learning*. Jmlr-Journal Machine Learning Research, Honolulu, HI.

Sitek, P., Wikarek, J., Jagodzinski, M., 2022. A Proactive Approach to Extended Vehicle Routing Problem with Drones (EVRPD). *Applied Sciences-Basel* 12(16), 21.

Sutskever, I., Vinyals, O., Le, Q., 2014. Sequence to Sequence Learning with Neural Networks, *28th Conference on Neural Information Processing Systems (NIPS)*. Neural Information Processing Systems (Nips), Montreal, CANADA.

Vaswani, A., Shazeer, N., Parmar, N., Uszkoreit, J., Jones, L., Gomez, A.N., Kaiser, L., Polosukhin, I., 2017. Attention Is All You Need, *31st Annual Conference on Neural Information Processing Systems (NIPS)*. Neural Information Processing Systems (Nips), Long Beach, CA.

Wang, X.Y., Poikonen, S., Golden, B., 2017. The vehicle routing problem with drones: several worst-case results. *Optimization Letters* 11(4), 679-697.

Wang, Y., Wang, Z., Hu, X.P., Xue, G.Q., Guan, X.Y., 2022. Truck-drone hybrid routing problem with time-dependent road travel time. *Transp. Res. Pt. C-Emerg. Technol.* 144, 27.

Wang, Y., Yang, X.X., Chen, Z.B., 2023. An Efficient Hybrid Graph Network Model for Traveling Salesman Problem with Drone. *Neural Process. Lett.* 55(8), 10353-10370.

Wu, G.H., Mao, N., Luo, Q.Z., Xu, B.J., Shi, J.M., Suganthan, P.N., 2022. Collaborative Truck-Drone

Routing for Contactless Parcel Delivery During the Epidemic. *Ieee Transactions on Intelligent Transportation Systems* 23(12), 25077-25091.

Xu, Y.Q., Fang, M., Chen, L., Xu, G.Y., Du, Y.L., Zhang, C.Q., 2022. Reinforcement Learning With Multiple Relational Attention for Solving Vehicle Routing Problems. *Ieee Transactions on Cybernetics* 52(10), 11107-11120.

Yu, V.F., Lin, S.W., Jodiawan, P., Lai, Y.C., 2023. Solving the Flying Sidekick Traveling Salesman Problem by a Simulated Annealing Heuristic. *Mathematics* 11(20), 21.

Zhang, K., He, F., Zhang, Z.C., Lin, X., Li, M., 2020. Multi-vehicle routing problems with soft time windows: A multi-agent reinforcement learning approach. *Transp. Res. Pt. C-Emerg. Technol.* 121, 14.

Zhang, S., Liu, S.L., Zhang, W.Y., 2023. Vehicle routing problems with time windows under the collaborative delivery mode of electric vehicle-drone. *Chinese Journal of Management Science*.

Zhao, J.X., Mao, M.J., Zhao, X., Zou, J.H., 2021. A Hybrid of Deep Reinforcement Learning and Local Search for the Vehicle Routing Problems. *Ieee Transactions on Intelligent Transportation Systems* 22(11), 7208-7218.

Zhou, J., Yu, Q., Xue, Z.M., Yang, L.B., 2025. Research on the Route Planning Problem of Drone and Truck Collaborative Delivery in Restricted Areas. *IEEE Access* 13, 33062-33073.

Zhuang, J.T., Tang, T., Ding, Y.F., Tatikonda, S., Dvornek, N., Papademetris, X., Duncan, J.S., 2020. AdaBelief Optimizer: Adapting Stepsizes by the Belief in Observed Gradients, *34th Conference on Neural Information Processing Systems (NeurIPS)*. Neural Information Processing Systems (Nips), Electr Network.